\documentclass[nonacm,  sigconf, screen]{acmart}
\usepackage{tikz}
\usepackage{standalone}
\usepackage{multirow}
\usepackage{enumitem}
\usepackage{graphicx}
\usepackage{xspace}
\usepackage{xcolor}
\usepackage{bbding}

\definecolor{purple}{HTML}{47289A}

\newcommand{\twozero}{\texttt{2\hspace{-1pt}.\hspace{-1pt}0}\xspace}
\newcommand*{\img}[1]{%
  \raisebox{-.22\baselineskip}{%
    \includegraphics[
      height=\baselineskip,
      keepaspectratio,
    ]{#1}%
  }%
}

\newcommand{\name}{\img{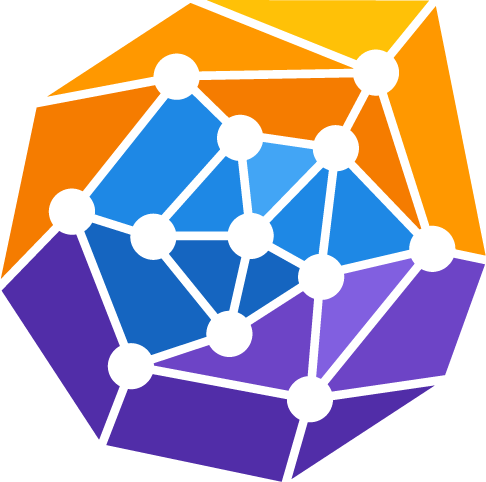}\,\textbf{Py\textcolor{purple}{G}~\twozero{}}}

\AtBeginDocument{%
  }

\setcopyright{acmlicensed}
\copyrightyear{2025}
\acmYear{2025}
\acmDOI{XXXXXXX.XXXXXXX}
\acmConference[TGL Workshop, KDD 2025]{Temporal Graph Learning Workshop, SIGKDD International Conference on Knowledge Discovery and Data Mining 2025}{August 03-07,
  2025}{Toronto, Canada}
\acmISBN{978-1-4503-XXXX-X/2018/06}
\begin{document}

\title{
\texorpdfstring{\name{}}{PyG 2.0}: Scalable Learning on Real World Graphs}

\author{
\fontsize{10}{12}\selectfont
\textbf{Matthias Fey}$^1$,
\textbf{Jinu Sunil}$^1$,
\textbf{Akihiro Nitta}$^1$,
\textbf{Rishi Puri}$^2$,
\textbf{Manan Shah}$^1$,
\textbf{Blaž Stojanovič}$^1$, \\
\textbf{Ramona Bendias}$^1$,
\textbf{Alexandria Barghi}$^2$,
\textbf{Vid Kocijan}$^1$,
\textbf{Zecheng Zhang}$^1$,
\textbf{Xinwei He}$^1$, \\
\textbf{Jan Eric Lenssen}$^{1,3}$,
\textbf{Jure Leskovec}$^{1,4}$\\[0.5em]
}

\affiliation{$^1$Kumo.ai  $^2$Nvidia  $^3$Max Planck Institute for Informatics $^4$Stanford University \\[0.5em]}
\renewcommand{\shortauthors}{Fey et al.}

\makeatletter
\def\@ACM@checkaffil{}
\makeatother

\begin{abstract}
\emph{PyG (PyTorch Geometric)} has evolved significantly since its initial release, establishing itself as a leading framework for Graph Neural Networks.
In this paper, we present \name{} (and its subsequent minor versions), a comprehensive update that introduces substantial improvements in scalability and real-world application capabilities. We detail the framework's enhanced architecture, including support for heterogeneous  and temporal graphs, scalable feature/graph stores, and various optimizations, enabling researchers and practitioners to tackle large-scale graph learning problems efficiently. Over the recent years, PyG has been supporting graph learning in a large variety of application areas, which we will summarize, while providing a deep dive into the important areas of relational deep learning and large language modeling.
\end{abstract}

\settopmatter{printfolios=true}

\maketitle

\section{Introduction}

Graph Neural Networks (GNNs) have emerged as powerful tools for learning on ubiquitous graph-structured data.
From social networks, knowledge bases, relational databases, to spatial graphs describing molecular structures, 3D scenes or objects, graphs are used to store most of the world's data.
Since 2019, \emph{PyG (PyTorch Geometric)}~\cite{fey2019fast} has been an important cornerstone in advancing deep learning on all these different types of graphs (\emph{cf.}~Sec.~\ref{sec:applications} for a summary).
PyG introduced a general message passing scheme that allows for a flexible formulation of Graph Neural Networks.
This is achieved by decomposing neural message passing~\cite{gilmer2017mpgnn} into \textsc{message}, \textsc{aggregation}, and \textsc{update} functions that can be customized to create various types of graph-based operators, thus supporting a broad range of models in a unified framework, which can automatically be mapped onto GPUs.

In the early years, most applied research around Graph Neural Networks revolved around finding the best operators to solve small-scale benchmark tasks, such as node classification on the Cora citation graph~\cite{Sen2008cora,Kipf/Welling/2017,Velickovic/etal/2018}, the graph-based equivalent to MNIST~\cite{deng2012mnist}.
Since then, the field of graph learning has rapidly evolved, strongly supported and driven by advancements in infrastructure provided by PyG.
GNNs can now be trained efficiently on web-scale, heterogeneous, temporal and multi-modal graphs, are explainable, and easily deployable for a wide range of practical applications.
PyG has evolved into a comprehensive blueprint for end-to-end graph-based machine learning, enabling these functionalities.

\begingroup
\renewcommand{\thefootnote}{}
\footnotetext{Published as a worshopt paper at KDD 2025}
\endgroup

In this work, we present the design principles and architectural decisions behind PyG, beginning with the foundational changes introduced in PyG 2.0 and extending through its continuous evolution to the current state of the library. PyG 2.0 marked a significant milestone in the library's development over three years ago, this paper encompasses the full trajectory of improvements and innovations that have been integrated into PyG up to its most recent version. We focus on the following three core aspects that have been refined and expanded throughout this evolution:
\begin{itemize}[label=\textcolor{purple}{\textbullet},leftmargin=10pt]
  \item \textbf{Heterogeneity.}
  Real world graphs have diverse node and edge types.
  PyG natively supports heterogeneous graph data types and message passing, as well as functionality for learning on temporal graphs.
  \item \textbf{Scaling and Efficiency.}
  Many use cases of graph learning have massive graphs ($\sim 10$ billion nodes), which need to be supported through optimized loading and training APIs.
  To this end, we present novel distributed processing capabilities, efficient data formats, loaders, and samplers, accelerated message passing, and compilation mechanisms.
  \item \textbf{Explainability.}
  Understanding how a model arrives at its decision is crucial in several domains and often required for trust in deep learning models deployed in practice.
  We discuss explainability in the heterogeneous graph learning setting and describe our plug-and-play method to make any GNN within PyG explainable out-of-the-box.
\end{itemize}

Graph learning powered by PyG has made an impact in a wide range of practical fields.
To showcase its generality, we also provide an overview of applications in chemistry, material design, computer vision, weather, and traffic forecasting. Moreover, we deep-dive into two specific application areas: GNN (and PyG) integration in Large Language Models~\citep{he2024gretriever} and Relational Deep Learning~\citep{fey2024rdl}. 

\section{\texorpdfstring{\name{}}{PyG 2.0}: End-to-End Graph Learning}

\begin{figure*}[t]
\centering
\includegraphics[width=\textwidth]{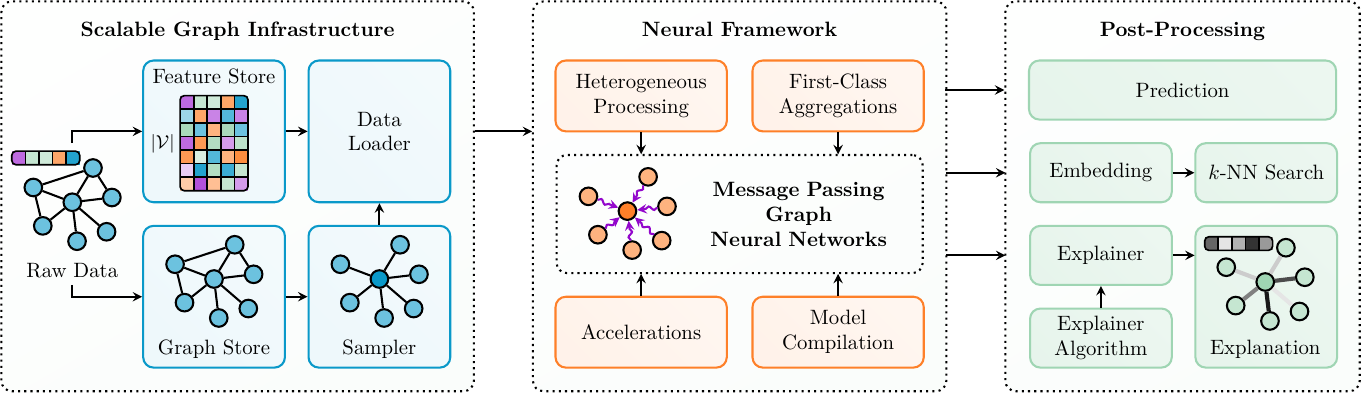}
\caption{Architectural overview of PyG~\twozero{}: The system's modular design allows to swap out any component without affecting other parts of the pipeline. For example, one can seamlessly change the \texttt{FeatureStore} from in-memory to distributed key-value storage without modifying the \texttt{DataLoader} or model parts. This plug-and-play approach extends throughout the framework---from storage implementations to sampling strategies and explainers. The core neural framework incorporates multiple performance optimizations including GPU accelerations, heterogeneous processing and model compilation techniques, ensuring efficient GNN training even on large-scale, heterogeneous and temporal graphs.}
\label{figure:overview}
\end{figure*}

In this section, we describe the building blocks that assemble the blueprint for end-to-end graph learning with PyG\footnote{PyG repo: https://github.com/pyg-team/pytorch\_geometric}.
We begin by providing an overview of all discussions in Sec.~\ref{sec:overview}.
Then, the subsequent sections outline the individual components in more detail, such as implementation aspects of our heterogeneous neural framework in Sec.~\ref{sec:neural}, scaling to real world graphs in Sec.~\ref{sec: scale}, and post-processing capabilities, \emph{e.g.}, via explainability, in Sec.~\ref{sec:explainability}.

\subsection{Framework Overview}
\label{sec:overview}

PyG is a library built upon \img{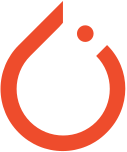}\,PyTorch~\citep{pytorch} to easily write and train Graph Neural Networks for a wide range of applications related to structured data.
It utilizes a tensor-centric API, \emph{i.e.} it exclusively operates on tensor-like data to define feature representations, graph structures and neural building blocks, and thus offers an intuitive experience which facilitates straightforward integration within the broader PyTorch ecosystem.
This design principle allows PyG to keep up-to-date with advances of its core, \emph{e.g.}, nested tensors\footnote{\texttt{torch.nested}: \url{https://pytorch.org/tutorials/prototype/nestedtensor}} for handling heterogeneous data of varying size, \texttt{TorchScript}\footnote{\texttt{TorchScript}: \url{https://pytorch.org/tutorials/beginner/Intro_to_TorchScript_tutorial}} for model serialization and deployment, \texttt{torch.fx}~\citep{torchfx} for model transformations, or \texttt{torch.compile}\footnote{\texttt{torch.compile}: \url{https://pytorch.org/tutorials/intermediate/torch_compile_tutorial}} for model optimization via Just-In-Time (JIT) compilation.
PyG leverages vectorized operations throughout its pipeline to maximize efficiency.
In cases where vectorization is not feasible—such as during graph sampling—we provide specialized C++ and CUDA kernels via our external (but optional) low-level \textbf{\texttt{pyg-lib}\footnote{https://github.com/pyg-team/pyg-lib}} package.

Figure~\ref{figure:overview} illustrates the comprehensive architecture of PyG, highlighting its modular and plug-and-play design.
The system is broadly organized into three main components: \textbf{(1)} graph infrastructure, \textbf{(2)} a neural framework, and \textbf{(3)} post-processing routines.
The graph infrastructure (\emph{cf.}~Sec.~\ref{sec: scale}) manages the lifecycle of (heterogeneous and temporal) graph data, supporting multi-modal feature processing, graph conversions, multi-threaded graph samplers, and distributed training.
The neural framework (\emph{cf.}~Sec.~\ref{sec:neural}) builds upon this data pipeline to define core interfaces and implementations for graph learning.
It offers efficient support for (heterogeneous) message passing, sparse aggregation operations, GPU acceleration, and model compilation.
Finally, post-processing (\emph{cf.}~Sec.~\ref{sec:explainability}) routines operate on the output of graph-based models to generate explanations, compute evaluation metrics, or perform $k$-nearest neighbor searches.

PyG's design offers flexibility via standardized interfaces throughout the full end-to-end pipeline.
One can easily swap components independently---transitioning from in-memory storage to databases, changing sampling strategies, or updating model architectures---all without disrupting other parts of the system.
The same interfaces work consistently whether we are handling small graphs or massive networks.
This architecture makes PyG particularly research-friendly, as it enables easy experimentation with novel techniques at any stage of the pipeline.

%Researchers can implement and test new storage systems, sampling methods, modeling approaches, or explainability tools by simply plugging them into the existing framework, allowing them to focus on innovation rather than rebuilding infrastructure for each experiment.

\subsection{Neural Framework}
\label{sec:neural}

\emph{Message Passing Graph Neural Networks (MP-GNNs)}~\citep{gilmer2017mpgnn, fey2019fast} are a generic framework to define a wide range of graph-based deep learning architectures.
Given a graph $G=(\mathcal{V}, \mathcal{E})$ with input node embeddings $\{\mathbf{h}^{(0)}_v\}_{v\in\mathcal{V}}$ and edge embeddings $\{\mathbf{e}_{(v,w)}\}_{(v,w)\in\mathcal{E}}$,
a single neural message passing step updates the node features by 
\begin{equation}
\label{eq:mp}
   \mathbf{h}^{(\ell+1)}_v = f \left( \mathbf{h}^{(\ell)}_v, \left\{\hspace{-0.1cm} \left\{ g \left( \mathbf{h}^{(\ell)}_w, \mathbf{e}_{(w,v)}, \mathbf{h}^{(\ell)}_v \right) \mid w \in \mathcal{N}(v) \right\} \hspace{-0.1cm} \right\} \right)\textnormal{,}
\end{equation}
where $f$ and $g$ are differentiable, optimizable functions and $\{\hspace{-0.1cm}\{\cdot\}\hspace{-0.1cm}\}$ a permutation invariant set aggregator, such as mean, max, sum. PyG automatically maps all implementations of the framework efficiently to GPUs by alternating between parallelization over edges (function $g$) and nodes (function $f$).
Almost all recently proposed GNN operators can be mapped to this interface, including (but not limited to) the methods already integrated into PyG~[\citealp{Kipf/Welling/2017,Velickovic/etal/2018,hamilton2017inductive,Defferrard/etal/2016,gilmer2017mpgnn}, and many others].

\paragraph{Accelerated Message Passing.}

% Three things:
% 1. Custom kernels (spmm, grouped/segment matmuls via CUTLASS)
%    * spmm
%      * paper: Design Principles for Sparse Matrix Multiplication on the GPU (hmmm it's a bit old; there must be faster spmm kernels out there)
%      * code: https://github.com/owensgroup/merge-spmm
%    * grouped/segment matmuls: https://pytorch-scatter.readthedocs.io/en/latest/index.html
% 2. CSC/CSR representation cache as an attribute of COO tensor subclass    
% 3. Kernel fusion via torch.compile (no graph breaks, no unnecessary recompilations)

As the primary operation in GNNs, message passing becomes a performance bottleneck, making its efficiency essential.
Within the first iteration of PyG, message passing was implemented by explicitly materializing $(\mathbf{h}^{(\ell)}_w, \mathbf{e}_{(w,v)}, \mathbf{h}^{(\ell)}_v)$ into edge-level space, followed by an aggregation into node-level space using atomic operations~\citep{fey2019fast}.
While easy to implement and effective to parallelize, memory requirements can become a bottleneck on denser graphs.

With PyG~\twozero{}, we introduce a new and unified way to accelerate message passing, leading to less memory-bottlenecked GNN workflows while preserving full backward compatibility.
In order to achieve this, we introduce the \texttt{EdgeIndex} tensor, which holds pair-wise source and destination node indices in sparse Coordinate Format (COO) of shape ${\{ 1, \ldots, |\mathcal{V}| \}}^{2 \times |\mathcal{E}|}$.
\texttt{EdgeIndex} sub-classes a general \texttt{torch.Tensor}, and thus preserves the ease-of-use of regular COO-based PyG workflows.
However, it can hold additional (meta)data, \emph{e.g.}, its sort order (if present) or whether edges are undirected.
Furthermore, it introduces a caching mechanism for fast conversion to Compressed Sparse Row (CSR) or Compressed Sparse Column (CSC) sparse formats.
Caches are filled based on demand, and are maintained and adjusted over its lifespan.
As a result, message passing in PyG can now rely on this (meta)data information to choose the optimal message passing computation path:
First, if the \texttt{EdgeIndex} is sorted by row or column, we can efficiently leverage sparse matrix multiplications (SpMMs) \cite{yang2018design} and segmented aggregations in GNN layers.
This ordering enhances data locality, reduces memory requirements, and enables greater parallelism on GPUs.
Second, for repeated GNN layer execution, caching the graph's CSC and CSR formats significantly reduces overhead during the backward pass.
Without this cache, computing the transposed adjacency matrix $\mathbf{A}^\top$—derived from the edge set $\mathcal{E}$—would be repeatedly required.
Finally, for undirected graphs where $\mathbf{A} = \mathbf{A}^\top$, caching the CSR format becomes unnecessary, further saving memory and computation.

\paragraph{Aggregations as a First-Class Principle.}

One of the most critical components of GNNs is the choice of the aggregation function.
It may account for symmetry, invariance~\citep{inductivebias,gdl}, and the expressive power~\citep{xu2018powerful,Corso/etal/2020} of GNNs to capture different types of properties of graphs.
Other works~\citep{hamilton2017inductive,deepergcn,You/etal/2020} empirically show that the choice of aggregation function is crucial to the performance of GNNs, and even utilize multiple aggregations~\citep{Corso/etal/2020,anisotropicgnn} or learnable aggregations~\citep{deepergcn} to obtain substantial improvements.

Inspired by this work, we made the concept of aggregation a first-class principle in PyG~\twozero{}, which allows users to easily plug-and-play with all kinds of aggregations---from simple ones (\emph{e.g.}, mean, max, sum) to advanced ones (\emph{e.g.}, median, variance, standard deviation), learnable ones~\citep{deepergcn}, and unconventional ones (\emph{e.g.}, via LSTMs~\citep{hamilton2017inductive} or equilibrium~\citep{equilibrium})---which can be also seamlessly stacked together~\citep{Corso/etal/2020,anisotropicgnn}.
Unifying the concept of aggregation helps us to perform optimization and specialized implementations in a single place, which can be utilized within both message passing and global readouts.

\paragraph{Heterogeneous Message Passing.}

PyG~\twozero{} introduces enhanced support for heterogeneous graphs, allowing seamless handling of multiple node and edge types.
This capability is essential for real-world applications where graphs naturally contain different types of entities and relationships.
%PyG~\twozero{} supports handling heterogeneous graphs with a dedicated \texttt{HeteroData} class.

Formally, a \emph{heterogeneous graph} is a graph $\mathcal{G}=(\mathcal{V}, \mathcal{E}, \phi, \psi)$, where each node $v\in \mathcal{V}$ and edge $e\in \mathcal{E}$ corresponds to a type $\phi(v): \mathcal{V}\rightarrow \mathcal{T}$  and $\psi(e): \mathcal{E} \rightarrow \mathcal{R}$, respectively.
Then, \emph{heterogeneous message passing}~\citep{schlichtkrull2018relational, hu2020hgt} is a \emph{nested} version of Eq.~(\ref{eq:mp}), adding an aggregation over all incoming edge types to learn distinct messages for each node type.
Heterogeneity is natively supported by PyG~\twozero{}.
It provides heterogeneous data types, transformations, graph samplers, and can automatically turn any message passing GNN into a heterogeneous variant.
This is achieved via a custom \texttt{torch.fx}~\citep{torchfx} transformation, which takes in a homogeneous GNN, replicates its GNN layers for every edge type in $\mathcal{R}$, and then transforms its computation graph to perform bipartite message passing over every edge type, followed by a custom aggregation to bundle messages pointing to the same destination node type.

The main challenge to efficiently implement heterogeneous GNNs lies in the varying number of nodes that belong to each node type $T \in \mathcal{T}$, \emph{i.e.} node features can be understood as a set $\{ \mathbf{H}^{(\ell)}_T \}_{T \in \mathcal{T}}$, $\mathbf{H}^{(\ell)}_T \in \mathbb{R}^{N_T \times F}$, where the number of nodes $N_T$ may vary for each node type.
In cases of dedicated heterogeneous GNN instantiations~\citep{hu2020hgt,han,heat}, PyG leverages \emph{grouped} and \emph{segmented} matrix multiplications to implement parallel projections across node/edge types efficiently.
Such re-occurring operation in heterogeneous message passing is defined as $\{ \mathbf{H}^{(\ell)}_T \mathbf{W}^{(\ell)}_T \}_{T \in \mathcal{T}}$ based on the three-dimensional weight tensor $\mathbf{W}^{(\ell)} \in \mathbb{R}^{|T| \times F \times F'}$, and requires both backward implementations w.r.t $\mathbf{H}^{(\ell)}_T$ and $\mathbf{W}^{(\ell)}$.
Internally, we implement both forward and backward passes using high-performance libraries such as CUTLASS~\citep{cutlass}.

\paragraph{Model Compilation.}

PyTorch's eager mode excels during the development and debugging phase of model design.
However, in production, performance—both in terms of speed and memory efficiency—becomes a top priority.
PyG~\twozero{} supports kernel fusion via \texttt{torch.compile}, enabling end-to-end compilation without graph breaks.
This allows multiple operations—including sparse computations and feature transformations—to be fused into a single, highly optimized kernel.
As a result, memory access and kernel launch overheads are minimized, making message passing significantly faster, especially for deeper or wider GNNs.
In order to support \texttt{torch.compile} within the irregular input workflows of PyG to full efficiency, we have revisited our entire code base to \textbf{(1)} avoid graph breaks and \textbf{(2)} remove any device synchronizations.
Our \texttt{MessagePassing} interface supports \texttt{torch.compile} out-of-the-box, without any user adjustments required.
On average, we observe 2--3$\times$ speedup in runtime while maintaining predictive accuracy, \emph{cf.}~Table~\ref{tab:efficiency}.

\paragraph{Graph Transformers.}

Aligned with recent advances in graph machine learning, PyG~\twozero{} integrates state-of-the-art \emph{Graph Transformer} architectures~\citep{graphgps,wu2024sgformersimplifyingempoweringtransformers,polynormer,shirzad2023exphormer,kim2022pure} into its package, applicable for learning on both many small graphs and single large graphs.
Positional encodings, which capture graph topology, can be computed either during pre-processing or dynamically at runtime.
These models are built on unified interfaces and seamlessly incorporate components from traditional GNN workflows.

% Results from https://github.com/pyg-team/pytorch_geometric/pull/10184
\begin{table}
 \centering
 \begin{tabular}{lrrrrr}
   \toprule
\multirow{2}{*}{\textbf{Run Mode}} & \multirow{2}{*}{\textbf{GIN}} & \textbf{Graph} & \textbf{Edge} & \multirow{2}{*}{\textbf{GCN}} & \multirow{2}{*}{\textbf{GAT}} \\
 &  & \textbf{SAGE} & \textbf{CNN} & & \\
\midrule
Eager
& 9.56 & 9.45 & 98.51 & 19.73 & 29.72 \\
\texttt{compile}
& \textbf{2.86} & \textbf{2.79} & \textbf{58.12} & \textbf{4.62} & \textbf{8.32} \\
%compile + Trim
%& \textbf{1.98} & \textbf{1.96} & \textbf{23.11} & 5.86 & \textbf{7.93} \\
 \bottomrule
\end{tabular}
 \caption{Forward and backward pass runtime in milliseconds across different GNN architectures. The baseline uses default eager mode without compilation. Compilation provides 2--3$\times$ speedup. %Compilation+trimming achieves 4--5$\times$ speedup.
 Benchmark protocols open-sourced at \href{https://github.com/pyg-team/pytorch_geometric/pull/10184}{\texttt{github.com/pyg-team/pytorch\_geometric}}.}
 \label{tab:efficiency}
 \vspace{-21pt}
\end{table}

\subsection{Scalable Graph Infrastructure}
\label{sec: scale}

%\begin{figure}[h]
%  \centering
%  \includegraphics[scale=.2]{images/fs_gs.png}
%  \caption{A depiction of the feature and graph store interface, and their role in a general dataloading paradigm for scalable GNNs.}
%  \Description{A depiction of the feature and graph store interface, and their role in a general dataloading paradigm for scalable GNNs.}
%  \label{fs_gs}
%\end{figure}

Real-world graphs come in various shapes and sizes, and there is a growing interest in scaling GNNs to graphs with billions of nodes and multi-thousand-dimensional features.
Such large-scale data is typically stored in external systems, \emph{e.g.}, embedded databases—with growing interest in using frameworks like PyG to support mini-batch GNN training directly on top of these storage platforms.

To meet this need, PyG~\twozero{} introduces new \texttt{FeatureStore} and \texttt{GraphStore} remote backend interfaces that enable seamless interoperability with custom storage, all while maintaining the familiarity of the PyG training loop and core PyTorch abstractions.

For large or distributed graphs, in which node features and edge indices are stored in custom locations, users are only required to implement the relevant methods within the remote backend interface; the rest of the training loop looks identical to an in-memory implementation, and any distributed communication required is handled transparently by PyG. 

Support for custom feature and graph storages in PyG~\twozero{} is enabled by defining a clear separation of concerns within the library.
Concretely, the data loading loop is segmented into three components: a \emph{feature store}, a \emph{graph store}, and a \emph{graph sampler}, \emph{cf.}~Figure~\ref{figure:overview}.
The data loader calls the graph sampler with a set of seed nodes, which performs graph sampling on the graph store and returns a set of subgraph structures.
The data loader subsequently requests the features of sampled nodes and edges from the feature store, and joins the features with the sampled subgraph to construct a PyG mini-batch object that can be directly used within its neural framework.
Users that define custom feature handling are only required to specify the implementation of the \texttt{get} operation on their feature backend, and users that define custom graph handling are required to specify how sampling is performed against their graph representation.
As a result, while the graph and feature stores can be independently partitioned, replicated, and stored in optimized formats, the training loop can operate oblivious to these details.
These abstractions are also foundational to the in-memory storage formats used in PyG.
Specifically, both \texttt{Data} and \texttt{HeteroData} classes in PyG inherit from the \texttt{FeatureStore} and \texttt{GraphStore} interfaces, providing a unified mechanism for retrieving mini-batches from any type of data storage throughout the whole code base.

\paragraph{Efficient Subgraph Sampling}
% Two things:
% 1. Slow Python (interpreter overhead and GIL)
% 2. Hierarchical neighbor sampling (aka trimming)
% TODO: Add a code block?
% TODO: Maybe explain how each mini-batch is structured, and how pyg utilises num sampled nodes, to trim unnecessary nodes from the batch in each layer.

Subgraph sampling~\citep{hamilton2017inductive,Chen/etal/2018,shadow,Markowitz/etal/2021,Addanki/etal/2021,Chen/etal/2018b,Zou/etal/2019,Huang/etal/2018,hu2020hgt,Chiang/etal/2019,Zeng/etal/2020,gnnautoscale} is a common technique used to scale graph learning to large graphs.
Instead of aggregating messages from all neighbors, only a subset of neighbors up to $k$-hops are sampled for each node of interest.
This reduces memory and computational cost, making mini-batch training feasible even on billion-scale graphs.

Despite its advantages, subgraph sampling can be inefficient if implemented naively.
Pure Python-based implementations suffer from interpreter overhead and are constrained by the Global Interpreter Lock (GIL).
% and a naive implementation would compute embeddings for nodes that do not influence the final output, which results in wasting computation and memory.
To mitigate these issues, PyG introduces a high-performance custom C++ homogeneous and heterogeneous subgraph sampling pipeline that supports multi-threading both across edge types and across data loader workers.
The underlying implementation is highly flexible to support different needs:
Users can seamlessly move between disjoint or intersecting subgraphs within a mini-batch, and can tune the output to be either directional or bi-directional (\emph{e.g.}, in order to implement deep GNNs on shallow subgraphs~\citep{shadow,Addanki/etal/2021}).

Unlike other GNN libraries that return layer-wise 1-hop subgraphs for neighbor sampling~\citep{Wang/etal/2019/dgl}, PyG produces a single multi-hop subgraph.
This design enables seamless transitions between full-batch and mini-batch training, supports interchangeable graph sampling strategies, and promotes a clean separation between model architecture and data loading, making the overall workflow more modular and flexible.

However, in some scenarios, this flexibility comes at the cost of performance, as the model cannot exploit special characteristics of the underlying data loading routine.
One such limitation is that a GNN trained on a Breadth First Search (BFS)-generated subgraph learns representations for \emph{all} nodes at \emph{all} depths of the network, although nodes sampled in later hops do not contribute to the representations of seed nodes in later GNN layers anymore, thus performing redundant computation.
To maximize efficiency, we introduce a layer-wise pruning mechanism which progressively trims the adjacency matrix of the returned subgraph.
This progressive trimming is done by simply slicing the adjacency and feature matrices according to the BFS ordering on-the-fly, ensuring zero-copying throughout the process. This approach, combined with model compilation, leads to a 4--5$\times$ speed up, \emph{cf.}~Table~\ref{tab:efficiency2}.

\begin{table}
 \centering
 \begin{tabular}{lcrrrrr}
   \toprule
\multicolumn{1}{c}{\textbf{Run}} & \multirow{2}{*}{\textbf{Trim}} & \multirow{2}{*}{\textbf{GIN}} & \textbf{Graph} & \textbf{Edge} & \multirow{2}{*}{\textbf{GCN}} & \multirow{2}{*}{\textbf{GAT}} \\
\multicolumn{1}{c}{\textbf{Mode}} & & & \textbf{SAGE} & \textbf{CNN} & & \\
\midrule
Eager & \XSolid
& 9.56 & 9.45 & 98.51 & 19.73 & 29.72 \\
Eager & \Checkmark
& 3.74 & 3.71 & 38.76 & 10.20 & 15.79 \\
\texttt{compile} & \XSolid
& 2.86 & 2.79 & 58.12 & \textbf{4.62} & 8.32 \\
\texttt{compile} & \Checkmark
& \textbf{1.98} & \textbf{1.96} & \textbf{23.11} & 5.86 & \textbf{7.93} \\
 \bottomrule
\end{tabular}
 \caption{Forward and backward pass runtime in milliseconds across different GNN architectures. The baseline uses default eager mode without compilation. With both compilation and trimming enabled, runtimes can be improved by 4--5$\times$.
 Benchmark protocols open-sourced at \href{https://github.com/pyg-team/pytorch_geometric/pull/10184}{\texttt{github.com/pyg-team/pytorch\_geometric}}.}
 \label{tab:efficiency2}
 \vspace{-21pt}
\end{table}

%\begin{figure}[h]
%  \centering
  % \includegraphics{images/trimming.pdf}
%  \includegraphics[scale=0.1]{images/trimming.jpeg}
%  \caption{As the GNN layer gets deeper, ther}
%  \Description{missing}
%  \label{trimming}
%\end{figure}

\paragraph{Temporal Subgraph Sampling.}

PyG~\twozero{} supports both \emph{temporal} homogeneous and heterogeneous graphs as part of its subsampling routines~\citep{wang2021temporalsampling,fey2024rdl}. 
Temporal subgraph sampling in PyG enables seamless traversal of dynamic graphs over time, allowing to extract (sub)graph snapshots at any point in time.

Given a seed node $v$ and a seed timestamp $t$, the resulting $k$-hop subgraph $G^{\leq t}_k[v]$ around node $v$ is constructed such that all included nodes and edges respect temporal constraints—specifically, they must have appeared at or before timestamp $t$.
This ensures the subgraph contains no future information, thereby preventing temporal leakage.
For node and edge types lacking timestamps (\emph{e.g.}, institutions or locations), sampling is performed without applying temporal constraints.

A variety of temporal sampling strategies are supported, including uniform sampling, sampling the most recent $k$ nodes or edges, and annealing-based strategies that gradually bias sampling toward more recent elements. Within each mini-batch, the sampled subgraphs are guaranteed to be disjoint, permitting different seed timestamps across samples while maintaining temporal consistency.

\paragraph{cuGraph Integration.}

Based on our \texttt{FeatureStore} and \texttt{Graph}-\texttt{Store} abstractions, we enabled end-to-end GPU-accelerated PyG workflows via cuGraph integration.
The cuGraph<>PyG extension, part of the NVIDIA RAPIDS~\citep{rapids} framework, is built upon \emph{cuGraph}~\citep{cugraph} for GPU-accelerated graph analytics and sampling, and \emph{WholeGraph}~\cite{wholegraph-part1,wholegraph-part2} for GPU-accelerated distributed tensor and embedding storage.
This enables 2x-8x data loading speed-ups with minimal code change, even for single-GPU workflows, and can be easily extended to multi-node multi-GPU setups.
All workflows benefit from a fast bulk sampling process on the GPU, which generates samples for as many batches as possible in parallel.
Then, during the feature fetching stage, WholeGraph allows features to be distributed across workers efficiently, which minimizes synchronization overhead, reduces memory transfers, and removes redundant data copies.
Through cuGraph<>PyG, it is possible to achieve linear scaling when stacking additional GPUs.

\subsection{Explainability}
\label{sec:explainability}

Explainability of machine learning models has become increasingly important for a range of reasons, including trust, regulatory compliance, security, and ease of debugging.
Unlike traditional machine learning models, GNNs operate over irregular and relational data structures, making their decision-making processes inherently more difficult to interpret—particularly when it comes to understanding both feature and structural influence.

\begin{figure}[t]
\centering
\includegraphics[width=\linewidth]{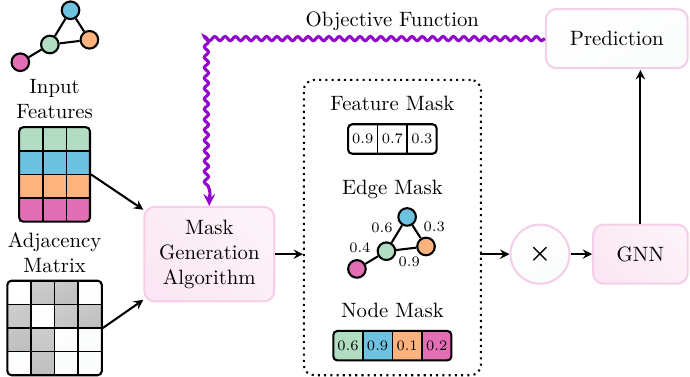}
\caption{Exemplary illustration of a GNN explainer in PyG: The explainer generates node-level and edge-level masks, which are multiplied within the GNN to weigh node features and message passing edges. The masks are optimized via an objective function to preserve only necessary information.}
\label{fig:explain}
\end{figure}

PyG~\twozero{} provides comprehensive support for explaining (heterogeneous) GNNs through its universal \texttt{Explainer} interface (\emph{cf.}~Figure~\ref{fig:explain}).
The \texttt{Explainer} class acts as a bridge between user-defined GNNs, explanation algorithms, and graph data, to generate attributions that signify the importance of nodes, edges, and features in the model's decision-making process.
Formally, given a $\mathrm{GNN}: \mathcal{G} \rightarrow \mathcal{Y}$, mapping an input graph to a prediction, we seek to find attributions $\mathbf{A}_{\mathcal{V}} \in \mathbb{R}^{|\mathcal{V}| \times F}$ and $\mathbf{a}_{\mathcal{E}} \in \mathbb{R}^{|\mathcal{E}|}$ that identify the contribution of an individual input feature in $\mathbf{H}^{(0)}$ and $\mathcal{E}$, respectively.

To generate structural explanations $\mathbf{a}_{\mathcal{E}}$ of non-differentiable inputs $\mathcal{E}$, the \texttt{Explainer} module temporarily alters the internal message passing process of Eq.~(\ref{eq:mp}) in PyG GNNs by enabling customization of messages through a callback mechanism $c: \mathbb{R}^{|\mathcal{E}| \times F} \rightarrow \mathbb{R}^{|\mathcal{E}| \times F}$, \emph{i.e.}
\begin{equation*}
   \mathbf{h}^{(\ell+1)}_v = f \left( \mathbf{h}^{(\ell)}_v, \left\{\hspace{-0.1cm} \left\{ c \left( g \left( \mathbf{h}^{(\ell)}_w, \mathbf{e}_{(w,v)}, \mathbf{h}^{(\ell)}_v \right) \right) \mid w \in \mathcal{N}(v) \right\} \hspace{-0.1cm} \right\} \right)\textnormal{.}
\end{equation*}
This callback allows, \emph{e.g.}, to introduce perturbations, to apply edge-level masks that weight incoming messages, or to capture internal attention coefficients.
Afterwards, the explanation algorithm assesses how these modifications affect the model’s predictions or align with ground-truth data~\citep{amara2022graphframex}.
Such callback mechanism is applicable both in homogeneous and heterogeneous GNNs.
In explanation mode, message passing falls back to edge-level materialization (c.f. Sec.~\ref{sec:neural}) in order to uniformly inject $c$ across all edges.

With this modular design, researchers only need to focus on the real challenge of building new and improved explainer algorithms, while the data flow, visualizations and evaluation protocols~\citep{agarwal2023evaluatingexplainabilitygraphneural,amara2022graphframex} (\emph{e.g.}, fidelity or unfaithfulness) are handled by the PyG framework.

\paragraph{Captum Integration.}

While PyG supports a variety of proposed graph-specific explainer modules~\citep{gnnexplainer,pgexplainer,graphmask}, it also provides a direct connection to \img{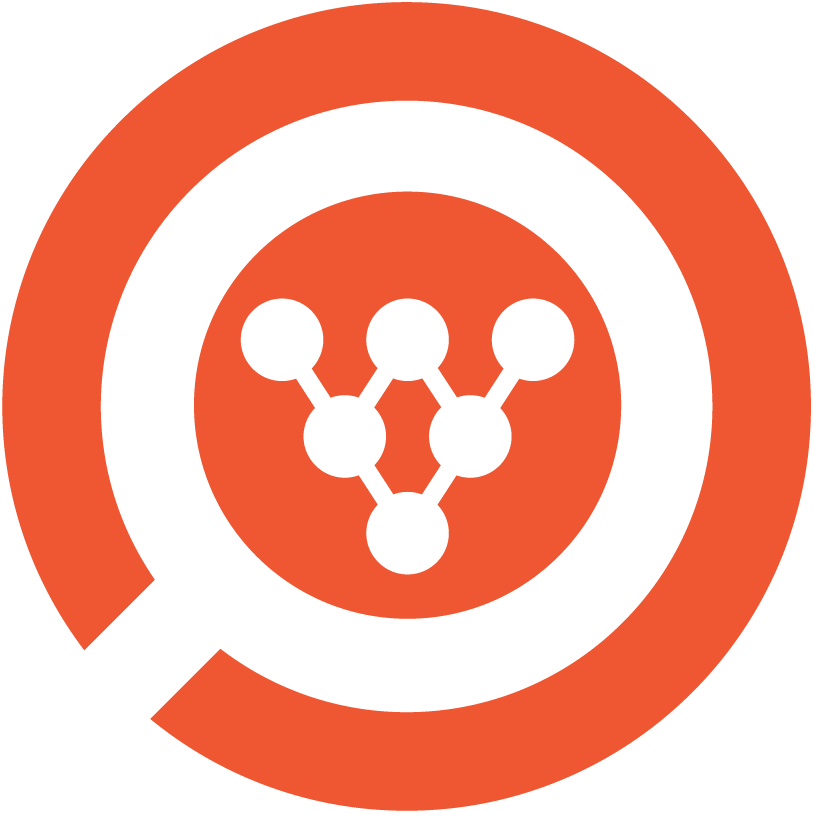}\,Captum~\citep{captum}, a general-purpose explainability library for PyTorch.
Captum offers a wide range of out-of-the-box explainers, such as saliency~\citep{saliency}, integrated gradients~\citep{ig}, guided backpropagation~\citep{guidedbp}, or deconvolution~\citep{deconvolution}.
While Captum is effective for various data modalities like vision and text, its direct application to GNNs presents challenges due to non-differentiable inputs $\mathcal{E}$.
As a consequence, our \texttt{CaptumExplainer} module builds a wrapper around any (heterogeneous) PyG GNN such that only the node features and an edge-level soft mask (initialized with ones) are required as input arguments.
Internally, the edge-level soft mask is then attached to reweigh messages in every GNN layer via the callback mechanism $c$.
This effectively makes all inputs to the GNN differentiable, which can now be utilized by Captum to explain both feature information and structural properties via its large set of gradient-based explainer modules.

\section{Applications: \texorpdfstring{\name{}}{PyG 2.0} in Action}
\label{sec:applications}

We now provide an overview of applications powered by PyG, with a special focus on Relational Deep Learning (Sec.~\ref{sec:rdl}) and integration in Large Language Models (Sec.~\ref{sec:llm}). Last, we provide a wider overview of further applications and the ecosystem (Sec.~\ref{sec:further_app}).

\subsection{Relational Deep Learning}
\label{sec:rdl}
PyG's support for heterogeneous temporal graphs enables its use for \emph{Relational Deep Learning (RDL)}~\citep{fey2024rdl}, offering a modern deep learning alternative to traditional feature-based approaches for learning on raw relational databases.
In RDL, relational data is represented as a graph, where each entity denotes a node, and the primary-foreign key links between entities define the edges.

\begin{figure}[t]
\includegraphics[width=0.8\linewidth]{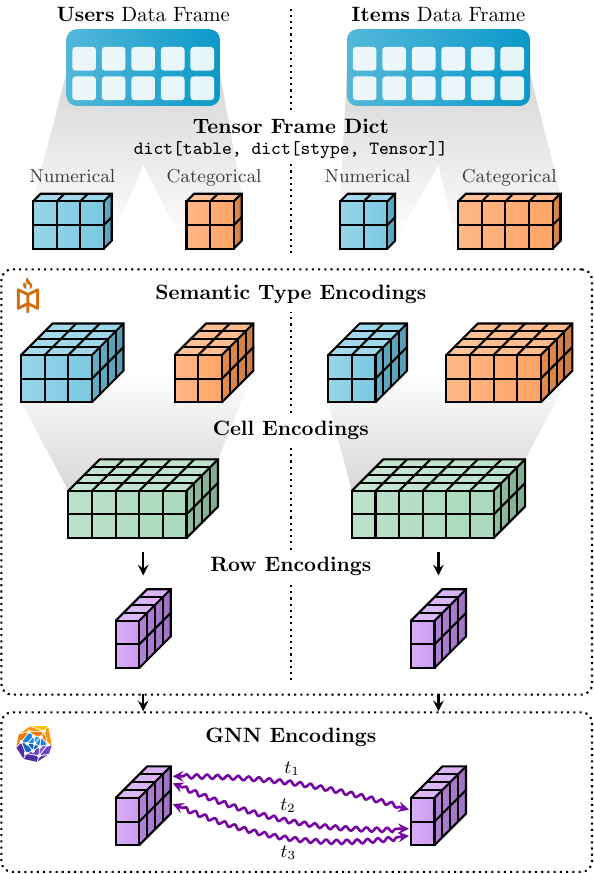}
\centering
\caption{End-to-end Relational Deep Learning on multi-modal and multi-table data with PyTorch Frame and PyG: Every row in each table is encoded individually using tabular deep learning algorithms. Afterwards, message passing GNNs can be applied to exchange cross-table information.}
\label{fig:rdl}
\end{figure}

PyG supports the full end-to-end RDL blueprint, which covers \textbf{(1)} handling of \emph{multi-modal} data, \textbf{(2)} querying historical subgraphs based on the contents of a \emph{training table}, and \textbf{(3)} \emph{recommender system} support.
For handling multi-modal data, we integrated support for \img{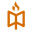}PyTorch Frame~\citep{hu2024pytorch} into feature fetching capabilities of our \texttt{FeatureStore}.
That is, we allow nodes to hold multi-modal data according to their semantic type (\emph{e.g.}, numericals, (multi-)categoricals, timestamps, free text), stored in a \texttt{TensorFrame}~\citep{hu2024pytorch}.
Afterwards, we can combine existing table-encoding algorithms~\citep{chen2023excelformer,gorishniy2021revisiting,chen2023trompt,huang2020tabtransformer,arik2021tabnet} from deep tabular learning jointly with GNN message passing algorithms for cross-table information exchange (\emph{cf.}~Figure~\ref{fig:rdl}).

Furthermore, RDL requires flexibility as part of data loading routines, where seed nodes, their timestamps, and corresponding labels are defined externally via a training table.
To accommodate this, PyG~\twozero{} enables subgraph samplers to iterate over externally specified seed nodes and timestamps, extracting subgraphs centered around the appropriate node types.
Ground-truth labels and other training table metadata can be dynamically attached to these subgraphs through the concept of \texttt{transforms}, which allow customization into the feature fetching pipeline.

Finally, PyG~\twozero{} offers full support for GNN-based recommender systems, including efficient \emph{Maximum Inner Product Search (MIPS)} via the FAISS library~\citep{faiss}, as well as mini-batch-compatible retrieval metrics (\emph{e.g.}, \texttt{map@k} or \texttt{ndcg@k}), implemented according to \texttt{torchmetrics}~\citep{torchmetrics} standards.
This elevates link prediction GNNs beyond the conventional binary classification paradigm—restricted to pre-defined candidate pairs—into realistic recommendation scenarios where candidate items are not known a priori.

%This creates a straightforward transformation of a relational database into a graph, allowing one to use GNNs directly on a database of multiple raw tables and skip the tedious manual feature engineering and aggregation into a single training table.
%Benchmarks such as RelBench~\cite{robinson2024relbench} demonstrate that classic tabular machine problems directly translate into known graph machine learning problems.
%Classification and regression problems become node classification and node regression, respectively, while recommendation translates to link prediction.

%To simplify the processing of relational databases, PyG integrates with PyTorch Frame library~\cite{hu2024pytorch}.
%This allows one to combine existing table-encoding algorithms from deep tabular learning jointly with the GNN message passing for cross-table message passing.
%As demonstrated by~\cite{robinson2024relbench}, this is sufficient to outperform decision tree based approaches while requiring significantly less manual input.

%RDL introduces \emph{training tables}, which can be automatically generated from a database, based on a SQL-like description. The training table is generically connected to the rest of the graph and defines a graph learning task by treating the corresponding nodes as target nodes. As a result, RDL can model even complex tasks, like recommendation, in an efficient way~\cite{yuan2024contextgnn}. 

\subsection{Integration in Large Language Models}
\label{sec:llm}

PyG contributes to the LLM domain in two different ways: \textbf{(1)} it provides examples how to utilize LLM embeddings as part of text-attributed graphs~\citep{tag1,harnessexpl,tag2} in graph learning, and \textbf{(2)} by supporting various techniques for \emph{Retrieval Augmented Generation (RAG)}~\cite{RAG}, as detailed in the following.

RAG enables LLMs to incorporate document databases as contextual knowledge sources.
To capture the underlying structure of these databases, models such as GNNs and Graph Transformers are used to enhance the LLMs' ability to reason over relational and topological information—a technique commonly referred to as \emph{GraphRAG}~\citep{msftgrag}.

%\subsubsection{What are RAG and GraphRAG}
%RAG\cite{RAG} is a popular technique for enabling an LLM to access a database of documents that it was not trained on in an effort to reduce hallucinations. The basic idea is to chunk up the documents and then use semantic similarities between a query and the chunks to provide the LLM with relevant context chunks for the query. This technique works well for simple queries but tends to struggle when the question requires information from across documents.

%This is where GraphRAG comes in. There are countless implementations of this but the one by Microsoft is one of the most popular\cite{msftgrag}. In general most of the approaches called GraphRAG are using LLMs for creating a knowledge graph from raw documents. They can then use LLMs to retrieve relevant information from this knowledge graph, even if that information originally spanned across documents.

%\subsubsection{Why GNNs are Important for GraphRAG}

\begin{figure}[t]
\includegraphics[width=\linewidth]{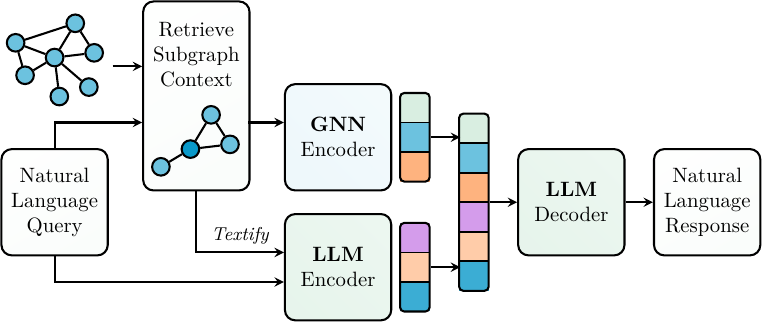}
\centering
\caption{The general GraphRAG pipeline in PyG:
A natural language query is used to retrieve relevant contextual subgraphs from a larger knowledge graph, which are then encoded via a GNN.
The resulting node embeddings are used to enhance the conventional LLM encoder<>decoder flow.}
\label{fig:rag}
\end{figure}

GraphRAG starts with a natural language query, which is used to retrieve a relevant contextual subgraph from a larger knowledge graph database.
This subgraph is then encoded using a GNN, and the resulting node embeddings are aggregated and projected into the LLM's embedding space.
PyG supports this retrieval workflow through extensions to its \texttt{FeatureStore} and \texttt{GraphStore} abstractions.
The interface is fully customizable and can be adapted to domain-specific retrieval strategies, \emph{cf.}~Figure~\ref{fig:rag}.

The full GNN+LLM generator pipeline in PyG is enabled via the \texttt{G-Retriever} model~\cite{he2024gretriever} which allows any combination of a PyG GNN with a HuggingFace LLM~\citep{huggingface}.
Notably, the addition of GNNs provides a 2x increase in accuracy over pure LLM baselines~\cite{neo4jblog}, improving from 16\% (LLM-based Agentic RAG) to 32\% (GNN+LLM-based Graph RAG) accuracy.

Furthermore, PyG provides the \texttt{TXT2KG} class, an easy-to-use interface to convert unstructured text datasets into a knowledge graph via parsing and prompt engineering.

\subsection{Others Application Areas}
\label{sec:further_app}
In recent years, PyG has been applied in a large variety of further application areas. In the following, we highlight a few important examples that showcase the wide range of applicability.

\paragraph{Chemistry.}
Graph neural networks and efficient GPU implementations such as PyG have been largely successful in chemistry~\cite{gao2023ppi, PandeyMohit2022Ttro}.
PyG has been used for drug discovery by combining GNN models with chemical foundation models~\cite{atz2024denovo}. Research infrastructure for material discovery with GNNs has been built on top of PyG~\cite{Fung2021BenchmarkingGN} opening applications of GNNs for surface material property prediction~\cite{Maurizi2022materials}.

\paragraph{Large Spatial Graphs.} Due to a large amount of data on connected nodes, PyG enables data-driven weather forecasting, as demonstrated by research in probabilistic weather forecasting~\cite{oskarsson2023graphbased, oskarsson2024probabilistic}. It was adopted by researchers from the European Centre for Medium-Range Weather Forecasts (ECMWF) to built a data-driven weather forecasting system~\cite{lang2024aifs}. Similarly, PyG was applied to analyze and predict behavior in traffic scenarios~\cite{klimke2022traffic, Huegle2020traffic}.

\paragraph{Optimization.} More recently, GNNs have become a dominant paradigm to solve combinatorical optimization problems~\cite{Schuetz2021CombinatorialOW, Cappart2023combinatorial}. Multiple solvers have been developed on top of PyG~\cite{karalias2020optimization, qiu2022optimization}. The field has been identified as one of the future fields with much potential GNNs development~\cite{bechlerspeicher2025positiongraphlearninglose}.

\paragraph{Social Network Analysis.} PyG has been used for a wide arrange of social network analysis tasks, such as bot detection~\cite{feng2022twibot, yang2023botdetection}, community detection~\cite{costa2023community}, and fake news detection~\cite{michail2022fakenews}.

\paragraph{Computer Vision.} In the area of computer vision, PyG has been applied to process irregularly structured data, such as unstructured point clouds~\cite{Zhang_2020_CVPR, Lenssen_2020_CVPR}, meshes~\cite{Fey_2018_CVPR}, and scene graphs~\cite{Zhang_2021_CVPR}. It found application to solve tasks like matching~\cite{Fey/etal/2020}, autonomous driving~\cite{yu2022autonomous}, and grasp analysis~\cite{Brahmbhatt2020grasp}.

\paragraph{Ecosystem.} PyG sparked the creation of a vibrant ecosystem of open-source software built on top and around it. Examples include Quiver~\cite{quiver2023}, a library for distributed training, AutoGL~\cite{guan2021autogl}, an AutoML framework,  DIG~\cite{liu2021dig} with higher level extensions, and Pytorch Geometric Temporal~\cite{rozemberczki2021pytorch} for temporal graphs. Additionally, PyGOD~\cite{liu2024pygod} adds functionality for outlier and anomaly detection, FedGraphNN~\cite{he2020fedmlresearchlibrarybenchmark} provides federated learning capabilities, and Pytorch Frame~\cite{hu2024pytorch} adds encoders for tabular data. In addition to functionality, benchmarks, such as Relbench~\cite{robinson2024relbench} for relational data and temporal graphs~\cite{huang2023temporal} have completed the package.

\section{Conclusion}
PyG~\twozero{} represents a significant advancement in graph learning frameworks, offering scalable solutions for real-world applications while maintaining ease of use and flexibility. We presented advances in three different categories, scalable graph infrastructure, the neural framework, and post-processing techniques such as explainability, showcasing the highly modular framework design. PyG has been applied in a wide range of applied fields, including the very recent areas of relational deep learning and RAG systems in large language models, in which we expect further significant developments in the near future.

\begin{acks}
Our deepest gratitude goes to the PyG open-source community for their invaluable contributions. We also thank our collaborators from NVIDIA---Serge Panev, Zachary Aristei, Junhao Shen, Rick Ratzel, Erik Welch, and Ralph Liu---as well as our partners at Intel for their support.
\end{acks}

\bibliographystyle{ACM-Reference-Format}
\bibliography{main}

%\appendix

\end{document}